\pdfoutput=1
\documentclass[11pt]{article}
\usepackage{amsmath}
\usepackage{graphicx}
\usepackage{natbib}
\usepackage{url}
\usepackage{color}

\begin{document}

\renewcommand{\mathbf} \vec

\markright{Multi-borders classification}

\title{Multi-borders classification}

\author{Peter Mills\\
Peteysoft Foundation\\
\textit{petey@peteysoft.org}
}

\maketitle

\begin{abstract}
The number of possible methods of generalizing binary classification to 
multi-class classification increases exponentially with the number of class labels.
Often, the best method of doing so will be highly problem dependent.
Here we present classification software in which the partitioning of multi-
class classification problems into binary classification problems is
specified using a recursive control language.

\end{abstract}


\section{Introduction}

Many statistical classification methods distinguish
between only two classes by drawing a hyper-surface in the feature space.
In a Support Vector Machine (SVM), the hyper-surface is drawn by minimizing
the classification error.  By implicitly transforming the feature space
through operations on the dot product, the shape of this hyper-surface can
be made quite complex \citep{Mueller_etal2001}.  In \citet{Mills2011} the
hyper-surface is discretely sampled by finding the root of the difference in 
conditional probabilities along a series of lines drawn between the two
classes.  The conditional probabilities are found using a kernel density
estimation technique \citep{Terrell_Scott1992} called Adaptive Gaussian
Filtering (AGF).

There are many methods of generalizing binary classification
schemes to more than than two classes.  
The LIBSVM library\citep{Chang_Lin2011}, 
for instance, uses a ``one-against-one'' approach wherein
each class is compared against every other class.
For large numbers of classes this approach is quite inefficient since 
there will be $n_c(n_c-1)/2$ binary classifications, where $n_c$ is
the number of classes.
Many other methods exist and the possible number
will increase exponentially with the number of classes.  

In many problems a different method of dividing or partitioning the
classes would be appropriate.  Consider four land surface types:
coniferous forest, deciduous forest, corn field and wheat field.
Here a hierarchical scheme (also called a decision tree)
seems most appropriate since the related
surface types will cluster together: first
discriminate between forest and field.  If forest is returned, then
discriminate between evergreens and hardwoods.  If field, then
between corn and wheat.  As another example, in a classification
problem involving discretized continuum values, it makes sense to
place the partitions between classes that define adjacent ranges
in the continuum data.

New extensions to the
{\it libAGF} library \citep{Mills2011} generalize the binary classification
problem so that the most appropriate method can be used to partition a 
multi-class problem without having to write a new program in each case.
The AGF borders-training method has been paired with this
algorithm, the combination of which we refer to as ``multi-borders''.
In what follows, we describe the rational behind the software, how it works 
and test it on an example problem comprised of discretized continuous data.  

\section{Theory}

\begin{figure}
\begin{center}
\includegraphics[width=0.9\textwidth]{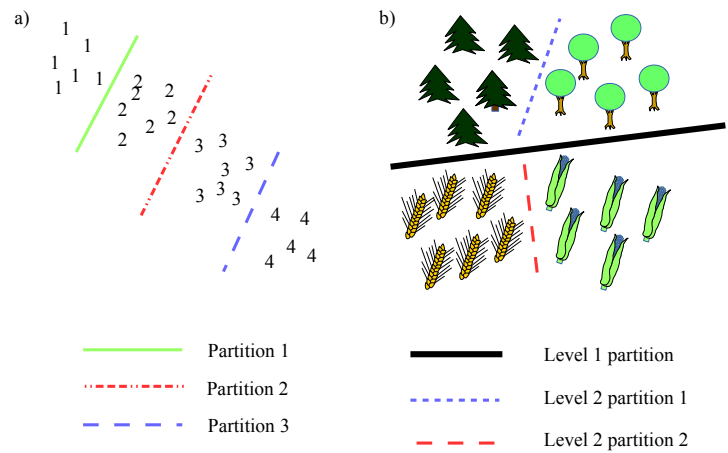}
\end{center}
\caption{Comparison between non-hierarchical (a)
and hierarchical classification (b).}
\label{figure}
\end{figure}

Suppose we have several partitions as in Figure \ref{figure}(a), each uniquely grouping
all the classes into two sets.
The following equations relate the conditional probabilities
of the classes to those returned by the binary partitions:
\begin{eqnarray*}
\sum_{j=1}^{n_c} P(j|\vec x) & = & 1\\
\sum_{j=1}^{n_{2i}} P(c_{2ij} | \vec x) - 
\sum_{j=1}^{n_{1i}} P(c_{1ij} | \vec x) & = & P_i(2 | \vec x) - P_i(1 | \vec x),
\end{eqnarray*}
where $P(c|\vec x)$ is the conditional probability of class $c$ at test point $\vec x$,
$P_i(1|\vec x)$ and $P_i(2|\vec x)$ are the conditional probabilities of the first and second
classes respectively on either side of partition $i$.  The classes contained
in either side of the $i$th partition are given by $\lbrace c_{1ij}|j=[1, n_{1i}]\rbrace$ and 
$\lbrace c_{2ij}|j=[1, n_{2i]}\rbrace$, respectively.

We call this {\it non-hierarchical} multi-borders classification.  
The popular ``one-against-the-rest'' approach, 
in which each class is singled out and classified against the remaining 
is one example of non-hierarchical classification 
and will be over-determined for every case.  
Note that the one-against-one approach is not covered by this
method nor by the hierarchical approach described below since it requires
excluding data from certain classes in absence of any prior knowledge of the 
class of the test point.

In a {\it hierarchical} classification scheme (or decision tree), the classes are first partitioned, then
each of those partitions are partitioned again and so on 
until a class number is returned instead of another partition.
The scenario for the first example is illustrated in Figure \ref{figure}(b).
A big difference between this and the non-hierarchical approach, is that all
data from classes in the losing partition are excluded from subsequent
classifications, whereas in the non-hierarchical approach all the data is
used in all the binary classifications.
As a corollary, in hierarchical multi-borders classification, 
only the conditional probability of the winning class is returned, 
whereas the non-hierarchical method solves for all of them.
The two types can of course be combined.

\section{Control language}

A recursive control language is used to describe any possible configuration in this hierarchical approach.  In Backus-Naur
form, the control language looks like this:

\begin{tabular}{lcl}
$<$branch$>$ & ::= & $<$model$>$ ``\{'' $<$branch-list$>$ ``\}'' $|$ $<$CLASS$>$\\
$<$model$>$  & ::= & $<$TWOCLASS$>$ $|$ $<$partition-list$>$\\
$<$branch-list$>$ & ::= & $<$branch$>$ $|$ $<$branch-list$>$ $<$branch$>$\\
$<$partition-list$>$ & ::= & $<$partition$>$ $|$ $<$partition-list$>$ $<$partition$>$\\
$<$partition$>$ & ::= & $<$TWOCLASS$>$ $<$class-list$>$ `` / '' $<$class-list$>$ ``;''\\
$<$class-list$>$ & ::= & $<$CLASS$>$ $|$ $<$class-list$>$ `` '' $<$CLASS$>$
\end{tabular}.

$<$CLASS$>$ is a class value between 0 and $n_c-1$.  It is used in two senses.
It may be one of the class values in a partition in a non-hierarchical model.
In this case it's value is relative, that is local to the non-hierarchical model.
It may also be the class value returned
from a top level partition in the hierarchy in which case it's value is absolute.

$<$TWOCLASS$>$ is a binary classification model.
There are two versions of control file: one for training and one for
classification using the trained model.  The command, \verb/multi_borders/,
reads a training control file and outputs a classification control file.
For training, $<$TWOCLASS$>$ contains a double-quoted set of parameters
or options for training a two-class model.  For classification, it 
is the name of a trained, binary classification model.
The \verb/multi_borders/ command returns a series of statements for training
each of the binary classifiers required for the over-all model, in addition
to the final control file which contains the names of each.

The \verb/classify_m/ command takes the output from \verb/multi_borders/
and uses it to perform classifications on a set of test data.
If the model has only one level, all the conditional probabilities are returned, 
otherwise only the winning probability is returned.
Command-line programs use AGF with borders sampling (\verb/class_borders/) 
as the binary classification model, however the source-level, C++ interface 
allows the user to specify any binary classification model desired.

\section{Test scenarios}

\begin{table}
\caption{Summary of multi-borders validation results}
  \begin{center}
     \begin{tabular}{|l|cccccc|} \hline
       Algorithm & train    & class.   & unc.   & acc. & corr. & correlation \\
                 & time (s) & time (s) & coeff. &      &       & cond. prob. \\ \hline\hline
AGF              & N/A & 235  & 0.43 & 0.56 & 0.92 & 1.   \\
Non-hierarchical & 189 & 2.0  & 0.41 & 0.53 & 0.91 & 0.94 \\
Hierarchical     & 111 & 0.84 & 0.42 & 0.54 & 0.91 & 0.89 \\
\hline
    \end{tabular}
  \end{center}
\label{multiclass_results}
\end{table}

To test the algorithm we use some of the same satellite humidity data as
described in \citet{Mills2009}.  
The specific humidity values are discretized into eight
classes by dividing them at seven geometricly increasing values from
0.001 to 0.00007.  Classes are labelled from 0 to 7 from lowest to highest
humidity ranges.  Two experiments were done.  The first used non-hierarchical
classification by partitioning the classes between each adjacent class, as
shown in the following control file:

\begin{verbatim}
"" 0 / 1 2 3 4 5 6 7;
"" 0 1 / 2 3 4 5 6 7;
"" 0 1 2 / 3 4 5 6 7;
"" 0 1 2 3 / 4 5 6 7;
"" 0 1 2 3 4 / 5 6 7;
"" 0 1 2 3 4 5 / 6 7;
"" 0 1 2 3 4 5 6 / 7;
{0 1 2 3 4 5 6 7}
\end{verbatim}

The blank options section means options can be passed from the command line.
The second experiment was hierarchical and partitioned the classes
recursively in half:
\begin{verbatim}
"-s 150 -W 40 -k 300" {
  "-s 100 -W 30 -k 250" {
    "-s 75 -W 25 -k 200" {0 1}
    "-s 75 -W 25 -k 200" {2 3}
  }
  "-s 100 -W 30 -k 250" {
    "-s 75 -W 25 -k 200" {4 5}
    "-s 75 -W 25 -k 200" {6 7}
  }
}
\end{verbatim}

The results from this experiment are shown in Table \ref{multiclass_results},
where they are compared with an AGF model with no pre-training.
While accuracy suffers somwhat using the multi-borders models,
there is an enormous improvement in classification speed, while
training times are less than the classification times for
the untrained model.

For the non-hierarchical model, conditional probabilities were solved
using a simple linear least squares.  Accuracy of estimates could likely
be improved by using constraints or regularization \citep{nr_inc2}.

\section{Conclusions}

Software was described that allows one to specify, in a recursive and
general way, a multi-class classification model comprised of one or more binary 
classifiers.  The system was tested on discretized satellite humidity data
using both a strictly hierarchical and strictly non-hierarchical model and
compared with a direct kernel estimator without any pre-training.
While the accuracy of both pre-trained models suffered somewhat compared to the
classifier without pre-training, time performance was greatly improved.
Software is available at: \url{http://libagf.sourceforge.net}.

\bibliography{multiborders_bib}

\end{document}